\documentclass[]{fairmeta}
\usepackage{lmodern}

\pdfoutput=1

\usepackage{latexsym}
\usepackage{amsmath}
\usepackage{adjustbox}
\usepackage{listings}
\usepackage{tikz}
\usepackage{pgfplots}
\pgfplotsset{compat=1.18}
\usepackage{xspace}
\usepackage{subcaption}
\usepackage{float}
\usepackage{tocloft}
\usepackage{booktabs}  
\usepackage{array}     
\usepackage{multicol}   
\usepackage[table,xcdraw,dvipsnames]{xcolor}

\usepackage{comment}
\usepackage{colortbl}
\usepackage{hyperref}
\usepackage{url}
\usepackage{booktabs}
\usepackage{xspace}
\usepackage{multirow}
\usepackage{colortbl}
\usepackage{booktabs}
\usepackage{multirow}
\usepackage{adjustbox} %
\usepackage{rotating}
\usepackage{graphicx}   
\usepackage{threeparttable}
\usepackage[textsize=tiny]{todonotes}
\usepackage{graphicx}
\usepackage{algorithm}
\usepackage{algpseudocode}
\usepackage{listings}
\usepackage{listings-ext}
\usepackage{wrapfig}
\usepackage{titletoc}
\usepackage{enumitem}

\newcommand{\sys}{\textsc{Refine-n-Judge}\xspace}

\newcommand{\new}[1]{\textcolor{black}{#1}}
\newcommand{\last}[1]{\textcolor{black}{#1}}

\title{\sys: Curating High-Quality Preference Chains for LLM-Fine-Tuning}

\author[1,*]{Derin Cayir}
\author[2]{Renjie Tao}
\author[2]{Rashi Rungta}
\author[2]{Kai Sun}
\author[2]{Sean Chen}
\author[2]{Haidar Khan}
\author[2]{Minseok Kim}
\author[2]{Julia Reinspach}
\author[2]{Yue Liu}

\affiliation[1]{Florida International University}
\affiliation[2]{Meta Reality Labs}

\contribution[*]{Work done at Meta}
\date{\today}

\correspondence{Derin Cayir at \email{dcayi001@fiu.edu}, Rashi Rungta at \email{rashirungta@meta.com}}

\abstract{Large Language Models (LLMs) have demonstrated remarkable progress through preference-based fine-tuning, which critically depends on the quality of the underlying training data. \last{While human feedback is essential for improving data quality, it is costly and does not scale well.} In this paper, we introduce \sys, an automated iterative approach that leverages a single LLM as both a refiner and a judge to enhance dataset quality. \last{Unlike existing iterative refinement methods, \sys uniquely employs an LLM to both generate refinements and explicitly evaluate each improvement, ensuring that every iteration meaningfully enhances the dataset without requiring additional human annotation or a separate reward model.} At each step, the LLM refines a response and judges whether the refinement is an improvement over the previous answer. This process continues iteratively \last{until the LLM prefers the initial answer over the refinement, indicating no further improvements.} This produces sequences of increasing quality, preference-labeled responses that are ideal for fine-tuning.

\last{We demonstrate the effectiveness of \sys across a diverse range of public datasets spanning five different corpora targeting various tasks such as coding, math, and conversation.} Models (Llama 3.1-8B and Llama 3.3-70B) fine-tuned on \sys-enhanced datasets were preferred by LLM judges in over 74\% of direct comparisons against models tuned on the original dataset \last{by GPT-4}. \last{Compared to prior approaches, our pipeline, guided by an effective judge, consistently produces valid refinements that improve dataset quality, making each iteration meaningful.} Additionally, we evaluated our models using established benchmarks, achieving notable performance gains (\sys models: increase of 5\% on AlpacaEval, and AlpacaEval 2.0, and 19\% increase on MT-Bench). Our results indicate that \sys effectively generates high-quality datasets that improve both the quality of fine-tuning datasets and capabilities of resulting models. This approach provides a scalable method for generating preference datasets and models in a continuous improvement loop. 
}

\begin{document}

\maketitle

\section{Introduction}

Large Language Models (LLMs) have shown remarkable performance across various tasks. However, \last{adjusting their outputs to match} human preferences remains a challenging and data-intensive problem. While fine-tuning LLMs on high-quality datasets can significantly enhance their performance on downstream tasks and improve alignment with user intent~\cite{rafailov2023direct, chung2024scaling}, the effectiveness of such fine-tuning is heavily dependent on the quality of the training data. 

Existing methods for curating datasets often involve human feedback. While beneficial, this approach is costly, limited in scalability, and sometimes unexplainable due to noise~\cite{rafailov2023direct}. Therefore, there is a need to focus on creating automated and scalable methods to improve the quality of training data~\cite{dubois2023alpacafarm}.

\begin{figure}
    \centering
    \includegraphics[width=0.6\columnwidth]{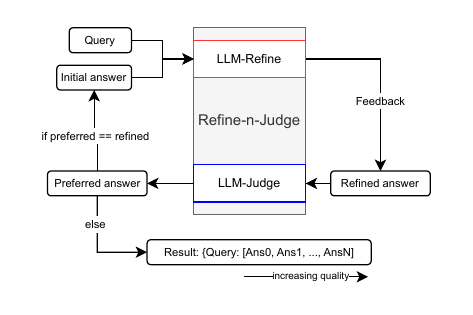}
    \caption{\last{The Refine-and-Judge process. Starting with a query and an initial answer, the LLM refines the answer based on generated feedback. It then compares the refined answer with the initial one and selects the preferred version. If the refined answer is preferred, it becomes the new initial answer; otherwise, the process ends.}}
    \label{fig:refine-and-judge-process}
\end{figure}

\new{A promising direction to improve data quality is to use LLMs for \textbf{refinement}. Previous research demonstrated LLMs' capablity of iteratively revising their own outputs to enhance aspects such as clarity, correctness, and informativeness, all without requiring human supervision~\cite{madaan2023self, novikov2025alphaevolve}.} However, due to the probabilistic nature of LLMs in generating outputs, not all refinements are guaranteed to be beneficial and what the model perceives as an improvement may not actually enhance quality.
\new{To address this limitation, a \textbf{judge} throughout the process is needed. LLMs have also shown to perform well as judges, achieving results comparable to human annotators when evaluating the quality of different responses~\cite{rafailov2023direct, dubois2023alpacafarm}. These models can assess nuances in response quality, with a consistency that often surpasses human evaluators. By leveraging their deep contextual understanding, LLMs provide objective and reproducible judgments that reduce variability and bias inherent in human annotation.}

By incorporating judgment into the refinement loop, we evaluate whether the newly generated response is truly better than the previous one. This not only improves the reliability of refinement, but also enables a fully automated and scalable framework for preference-based dataset curation, where refinements proceed only when quality gains are confirmed. 

\begin{figure}
    \centering
    \includegraphics[width=0.6\linewidth]{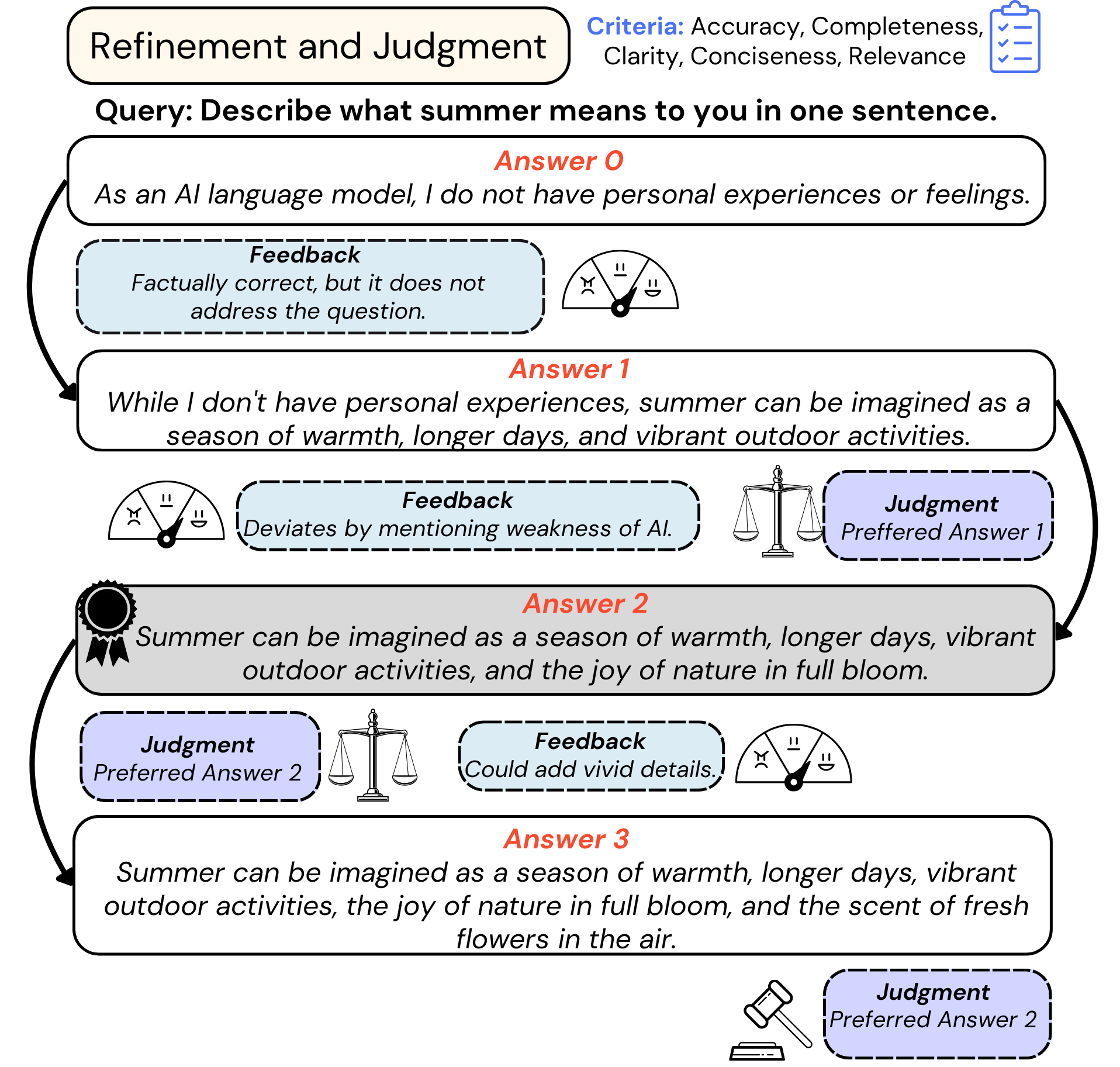}
    \caption{Example of \sys. Beginning with an initial model-generated output, each successive response is produced by prompting the model to refine the previous one, and the final selected response is highlighted in gray.}
    \label{fig:refine-and-judge-example}
\end{figure}

\last{Bringing these capabilities together, we propose \sys, a fully automated dataset curation pipeline, summarized in Figure~\ref{fig:refine-and-judge-example}.} In this framework, an LLM model serves as both the \textbf{refiner}- generating improved outputs- and the \textbf{judge}-comparing the refined output against the original and selecting the preferred version.  If the refined answer is preferred, the process continues with this new output as the next candidate for improvement. \last{However, if the judge perceives the refined answer to be worse than the initial answer, the process terminates by retaining the original answer. This loop enables scalable, iterative improvement to dataset quality without human intervention, resulting in a dataset that is a query and a list of answers in increasing quality.}

\new{We illustrate this interplay between refinement and judgment is illustrated in Figure~\ref{fig:refine-and-judge-example}, using an example from the TULU dataset. The refinement and judgment criteria included accuracy, completeness, clarity, conciseness, and relevance. We compared \sys to a pipeline that solely relies on iterative refinements without a judge (Figure \ref{fig:refinement_only_ex} in Appendix). Although both methods used the same criteria for the LLMs, these two figures demonstrated that unguided refinement does not perform as well as \sys. Incorporating judgment at each step ensured that only meaningful improvements are accepted. In \sys, the process terminated when the judge observed no further gains, as illustrated by the selection of the second answer over the third. However, without a judge, refinements continued even if they did not add value, as seen in Figure~\ref{fig:refinement_only_ex} where verbosity increases with continuous refinements. Therefore, by incorporating a judge, the model's answers are prevented from drifting through unnecessary continuous refinements.}

To analyze our proposed framework, we took responses to queries from publicly available benchmarks. Then, we applied our framework to enhance the dataset quality across \last{different generation tasks}, including code generation, mathematics, general question and answers, and natural language. Our proposed \sys framework outperformed baseline methods like zero-shot generation, and rejection sampling, \last{where GPT-4 prefers \sys 74\% of the time compared to refinement-only pipelines}. 

To further evaluate the quality of curated datasets, we fine-tuned Llama models using outputs generated by our pipeline. With Supervised Fine-Tuning (SFT), Llama 3.1-8B model fine-tuned on our curated dataset achieved 70\% win-rate compared to the same model fine-tuned on the original unrefined public data, as judged by GPT-4. Similarly, a larger Llama 3.3-70B model achieved an 75\% win rate under the same evaluation setup. %
These results suggested that preference based datasets curated from \sys enhanced model performance and \sys is an innovative way of curating high-quality datasets with preferences through LLMs' capabilities of refinement and judgment. Our results demonstrated that LLMs' judgment and refinement capabilities complement each other and provide a robust pipeline when combined. The curated dataset from this approach is a strong resource for fine-tuning LLMs, as our results demonstrate.

In summary, we make the following contributions:
\begin{itemize}
    \item We introduce \sys, a high-quality dataset curation method that utilizes LLMs' capabilities of refinement and judgment, offering a scalable, human-free approach to preference data collection, improving upon prior work.
    \item We applied \sys to a \last{five different corpora}, measuring the improvement in dataset quality, and used the curated datasets to fine-tune LLMs, and improved models across various benchmarks.
    \item \new{We created a large dataset with over 78,000 queries of preference-based ranked output chains using our pipeline, that we performed SFT.}
\end{itemize}

\section{Related Work}
Previous work has explored how LLMs can be used as a tool for refinement and judgment. \last{In this section, we summarize those works and our differences. A discussion on broader related work is given in the Appendix.}

\noindent\textbf{LLMs as Refiners:} Recent studies have investigated the potential of LLMs to iteratively refine their own outputs through self-generated feedback. \last{For instance, in SELF-REFINE framework~\cite{madaan2023self},} the model generates an initial output ($Answer_0$) that is subsequently revised based on feedback it has itself produced. \last{In this work, the model’s improvement is highly dependent on the quality of the feedback, where the feedback should be both actionable and specific for the model to be generating an improved output. However, due to this dependency on feedback quality, there is no guarantee that refinements will always lead to improved quality. In fact, refinements can sometimes degrade the output, given the probabilistic nature of LLMs in generating responses.}

\noindent\textbf{LLMs as Judges:} Recent studies have demonstrated the potential of LLMs to replace human-based evaluations by acting as self-judges~\cite{zheng2023judging, dubois2023alpacafarm}. These studies demonstrated that LLMs can approximate human preferences with reasonable accuracy, offering an appealing method to evaluate model outputs. However, despite their effectiveness, LLMs are prone to position bias, where the model may favor outputs based on their relative position, verbosity bias, where the model tends to select longer responses regardless of their quality, and self-enhancement bias, where the model consistently chooses its own output over other model-generated answers~\cite{zheng2023judging}. \last{Therefore, when using LLMs as judges, it is important to minimize these biases—for example, by swapping the positions of choices, and penalizing lengthy responses, a technique employed by \sys.}

\noindent\textbf{Using LLMs for Data Curation:} Another promising line of research has explored the use of LLMs for curating training datasets, particularly those augmented with AI-generated feedback~\cite{cui2023ultrafeedback}. For example, researchers created datasets where responses were generated by multiple models, and GPT-4 was tasked with annotating these responses with preferences and feedback. Additionally, other works have investigated using LLMs to synthesize datasets for text-based multi-agent simulations~\cite{tang2024synthesizing}. These synthetic datasets offer valuable resources for training and evaluating various NLP models. \last{Unlike these approaches, our proposed pipeline focuses on iteratively refining a single answer until no further improvements can be made, resulting in quality-ranked chains without relying on multiple LLMs to generate different answers or a separate reward model.}

\new{Despite these works, to the best of our knowledge, no prior work has fully explored a combined approach that leverages the dual capabilities of LLMs in refinement and judgment to enhance each other's performance. Moreover, integrating this pipeline to curate preference-based datasets remains an open research direction worthy of further investigation, which in this work we explore.}

\section{Approach}
Our proposed method is a \last{two-phase} approach (1) dataset method, and (2) training method. 

\subsection{Dataset Curation with \sys}
In the first part of the proposed method, we perform dataset curation from the iterative Refine-and-Judge process. %
\last{For a given query, an initial answer \new{could either be} obtained from publicly available datasets \new{or it could be generated by an LLM}.} Recognizing that this answer may not be of optimal quality, we employ an iterative process to synthesize a higher-quality dataset from the initial dataset. This iterative process consists of the following steps:

\noindent\textbf{1) Refinement:}
The process begins by obtaining the initial answer from a dataset (response to the query such as a question), \new{referred as $Ans_0$}. \new{The initial answer could also be generated from an LLM.}
Since this answer may not be the most ideal output for the query, we aim to iteratively refine $Ans_0$ into improved versions $Ans_1, Ans_2,...,Ans_n$ such that $Ans_n$ will become the most refined and highest quality output. \new{Here, the LLM will be presented both the question and the prior answer $Ans_t$ to generate the refined answer, $Ans_{t+1}$.}

\noindent\textbf{2) Judgment:}
Due to the probabilistic nature of LLM outputs, $Ans_{t+1}$ is not guaranteed to be superior to $Ans_t$. Therefore, the quality of refinements and whether they are improving \new{should be verified through an LLM}. Hence, we propose using LLM as a judge. \footnote{\last{While we use the same LLM (e.g., Llama 3.3-70B) for both refinement and judgment, different models can also be employed.}} LLM-Judge evaluates $Ans_t$ and $Ans_{t+1}$ to determine higher quality response \last{based on analyzing the responses on a set of criteria such as accuracy, completeness, clarity, conciseness, and relevance.}

\noindent\textbf{3) Iteration:}
This process iteratively continues until when the judge does not prefer the new response, indicating that the refined response is worse or around the same quality as the prior response. However, if the refined output $Ans_{t+1}$ is judged to be better then the prior response $Ans_t$, then $Ans_{t+1}$ becomes the new baseline. \last{The process repeats until termination. In cases where refinements continuously improve the output without the judge favoring the initial answer, we set a maximum of ten iterations to prevent indefinite refinement.}

This iterative refinement and judgment cycle results in a sequence of answers for each query: $[Ans_0, Ans_1, ..., Ans_n]$, where each subsequent version ($Ans_k$) is of demonstrably higher quality than its predecessor ($Ans_{k-1}$). 

\new{The iterative refinement and judgment process in \sys employs structured prompts designed for precise feedback, systematic answer refinement, and judgment. These prompts are given in Appendix. The pipeline consists of three sequential prompts (1) feedback prompt, used to evaluate the initial answers based on specific quality criteria, (2) refinement prompt, where feedback is incorporated to improve the answers, (3) judgment prompt, where the refined answers are compared against previous versions to select the superior response.}

\subsection{Fine-Tuning through Preference Chains}
The curated dataset $Query: [Ans_0, \ldots, Ans_n]$, which contains preference-ordered list of responses, can be used to fine-tune LLMs. 
\new{We fine-tune a pre-trained model using the final, highest-quality answers obtained from \sys. Specifically, we pair each query with its best refined answer, $Ans_n$ given the corresponding query. This allows the model to learn from the highest-quality outputs as determined by the pipeline.}

\section{Experimental Setup}

\subsection{Dataset and Metrics}
For the analysis of \sys, we focused on diverse datasets on assistant-style conversations in English. 

\new{We utilized these datasets for distinct purposes within our pipeline. The datasets, their respective purposes, and the evaluation metrics used are summarized in Table~\ref{tab:dataset_summary}. As shown in the table, some datasets were used to analyze specific aspects, such as robustness to various types of noise, while others were employed for fine-tuning the model. We examined the pipeline's robustness to inaccuracy, verbosity, misleadingness, and unhelpfulness, using evaluation metrics like accuracy, length, and external LLM assessments. To assess the effectiveness of our dataset curation method within the \sys pipeline, we fine-tuned LLMs and evaluated them using MT-Bench and AlpacaEval 2.0 benchmarks.}

\begin{table}[h]
    \centering
    \resizebox{0.6\columnwidth}{!}{%
        \begin{tabular}{lcc}
            \toprule
            \textbf{Dataset} & \textbf{Target Aspect} & \textbf{Evaluation Metric} \\
            \midrule
            Acronym & Inaccuracy & Accuracy \\
            UltraChat & Verbosity & Length \\
            TruthfulQA &  Misleadingness & LLM-as-a-judge score (0-100) \\
            OpenAssistant &  Helpfulness & LLM-as-a-judge score (0-100 \\
            TULU3 &  - & AlpacaEval 2.0, MT-Bench \\
            \bottomrule
        \end{tabular}
    }
    \caption{Datasets and their purposes in the pipeline. \last{Unless otherwise specified, GPT-4 is used as the LLM-Judge.}}
    \label{tab:dataset_summary}
\end{table}

\subsection{Pipeline Analysis Method}
We conducted experiments to demonstrate \sys's effectiveness by analyzing its core components:
\begin{itemize}
    \item \textbf{Iterative refinement:} To showcase the significance of continuous refinement, we compared \sys against rejection sampling, and one-shot answers produced by LLMs.
    \item \textbf{Judgment control:} To evaluate the importance of judgment within our pipeline, we compared \sys with a pipeline employing only repeated self-refinements~\cite{madaan2023self}. \new{With this, we showed the importance of having a judge in the pipeline compared to doing refinements without a stopping condition.}
\end{itemize}

Additionally, \new{recognizing that the datasets intended to be used with \sys may contain low-quality answers, we assessed the robustness of \sys to noise which may come from poor datasets. We set these answers as $Ans_0$ and applied our pipeline to convert them into high-quality responses. For this purpose, we created a custom dataset that targets inaccuracy, verbosity, and misleadingness through a combination between online datasets (e.g., TruthfulQA~\cite{lin2021truthfulqa}) and synthetic ones.} From TruthfulQA, we sampled one of the available multiple ``wrong answers" which answer a given question through misconceptions such as to ``What happens to you if you eat watermelon seeds?", the wrong answer that will be taken as $Ans_0$ is ``You grow watermelons in your stomach". We then compared the most-refined answer derived from the intentionally misleading $Ans_0$ to a true answer provided from the same dataset, using LLM judges. 

To target inaccuracy, we used Acronym dataset~\cite{madaan2023self} where the objective is to find an acronym for a given title. We generated inaccurate $Ans_0$ by using GPT-4 to create prompts that intentionally produce inaccurate yet believable answers. For verbosity, we used a dialogue dataset, Ultrachat~\cite{ding2023enhancing} where we obtained the questions and produce verbose, unhelpful responses through GPT4. 

\subsection{Fine-tuning Setup}
For fine-tuning LLMs, we used TULU~\cite{lambert2024tulu3}. 
\new{The TULU dataset is a high-quality instruction-tuning corpus that comprises diverse and extensive data collected from various domains, including academic question-answering, coding problems, reasoning tasks, and multilingual instructions. Due to its diversity and comprehensiveness, TULU served as a robust baseline for comparative analysis.}

\new{In our experiments, we first fine-tuned LLMs using the original TULU dataset, which provides a baseline performance. Then, we fine-tuned the same models using the refined dataset produced by \sys. This approach allowed us to measure the performance gains directly attributed from the pipeline. Specifically, we evaluated whether models fine-tuned on refined responses from \sys ($Ans_n$) exhibited improved performance over those trained solely on the original responses ($Ans_0$).} 

We applied this fine-tuning methodology to Llama 3.3-70B-Instruct, and Llama 3.1-8B-Instruct. \last{We used a learning rate of 1e-5 with a warmup phase of 200 steps, and the models were trained for a total of 11,000 steps. The batch size was set to 1, and the sequence length was extended to 32,768 to accommodate longer inputs. Model parallelism was employed with a size of 8 to efficiently distribute the computational load.} The performance of these fine-tuned models were evaluated across \last{AlpacaEval, AlpacaEval 2.0, and MT-Bench. More details are provided in Appendix.}

\section{Experiments}
\new{In this section we present our experimental findings where we first discuss the pipeline's details such as the need for continuous refinement, and judge. Then we analyze the pipeline's robustness to noisy data and the judge's consistency when making decisions. Lastly we present the results obtained from the fine-tuning stage to prove the curated dataset method's importance through \sys.}

\subsection{Importance of Continuous Refinement}

\new{Continuous refinement is computationally more expensive as it requires repeated calls to LLMs to obtain judgments and refinements. However, we found this cost (as detailed in Appendix~\ref{sec:gpu_appendix}) to be justified, as the performance of \sys was superior to other techniques.}

First, we analyzed how \sys compares to generating multiple answers in zero shot and selecting the best. Specifically, we used the same LLM for \sys to generate ten different answers to a given query in zero shot. Then using a separate prompt, we selected the best among the ten responses. We compared the best answer to the $Answer_n$, which was obtained as the ``most refined'' answer from \sys. We used GPT-4 as a judge to compare \sys with the method of generating multiple answers and selecting the best. With this approach, \sys (average iteration of 3.4) was preferred 98.4\% of the time by GPT-4 over generating 10 answers and selecting the best. This demonstrated that continuous iterations of \sys outperform the same model generating multiple answers and selecting the best one, highlighting the importance of a pipeline that incorporates refinements through a judge.

\begin{table}[ht]
    \centering
    \resizebox{0.4\columnwidth}{!}{%
    \begin{tabular}{lcc}
        \toprule
        \textbf{Dataset} & \textbf{Method} & \textbf{\% Wins} \\
        \midrule
        \multirow{3}{*}{Acronym} & \textbf{\sys} & \textbf{81.2} \\
                                 & Zero-shot Gen & 74.8 \\
                                 & Simple rule based & 62.4 \\
        \midrule
        \multirow{2}{*}{TruthfulQA} & \textbf{\sys} & \textbf{89.0} \\
                                    & Zero-shot Gen & 11.0 \\
        \midrule
        \multirow{2}{*}{OpenAssistant} & \textbf{\sys} & \textbf{84.8} \\
                                       & Zero-shot Gen & 15.2 \\
        \midrule
        \multirow{2}{*}{UltraChat} & \textbf{\sys} & \textbf{87.3} \\
                                    & Zero-shot Gen & 12.7 \\
        \bottomrule
    \end{tabular}
    }
    \caption{\sys performance comparison across different datasets: Acronym~\cite{madaan2023self}, OpenAssistant~\cite{kopf2023openassistant}, TruthfulQA~\cite{lin2021truthfulqa}, and UltraChat~\cite{ding2023enhancing}. For Acronym, we measure \% wins through accuracy (true or false acronym), others we used GPT-4 as a judge.}
    \label{tab:one_shot_vs_rj}
\end{table}

Second, we investigated whether iterative refinements are superior to one-shot answer generation. To do this, we obtained an initial answer $Answer_0$ to query from various datasets and then used \sys to generate $Answer_n$ from $Answer_0$. We also generated $Answer_{llm}$, a one-shot answer to the same query. We used GPT-4 as a judge to choose between $Answer_n$ and $Answer_{llm}$. Our results are presented in Table~\ref{tab:one_shot_vs_rj}.

As seen in Table~\ref{tab:one_shot_vs_rj}, \sys was preferred across all datasets. Incorporating iterative refinements and judgments in the pipeline yielded better results than generating answers in a one-shot manner using the LLM. Specifically, GPT-4 preferred answers improved by \sys over those generated from scratch. \last{This effect was particularly evident in the Acronym dataset, where the objective was to accurately determine the acronym of a title. A simple rule-based function, which generates acronyms from the first character of each word, achieved a 62.4\% success rate.} Generating one-shot answers to the given titles resulted in 74.8\% accuracy, but using \sys and refining the initial responses was more effective, resulting in an increase in accuracy to 81.2\%. These results suggested that the continuous refinement process employed by our pipeline is more effective than generating new answers from scratch via one-shot LLM outputs.

\subsection{Importance of LLM-Judge}
In \sys pipeline, LLM-judge is incorporated to ensure that each refinement made by the refiner can be validated before the pipeline continues with its iterations. This validated that the ``improvements" are meaningful and that the pipeline did not suffer from questionable refinements that reduced the quality of the output.

\begin{figure}[h]
  \centering
  \begin{tikzpicture}
    \begin{axis}[
      scale=0.4, %
      xlabel= Number of Refinements,
      ylabel=\parbox{3cm}{Win \% - RnJ over \\ Refinement-only},
      ymin=45, ymax=75,
      xtick={1, 2, 3, 4, 5, 6, 7, 8, 9, 10},
      ytick={50, 60, 70},
      legend pos=north west,
      enlargelimits=false, 
      width=\columnwidth,
      height=0.5\columnwidth,
    ]
      \addplot[
        color=blue, 
        mark=*,
        error bars/.cd,
        y dir=both, %
        y explicit %
      ]  coordinates {
        (2, 53) +- (0, 2.6)
        (3, 55) +- (0, 3.0)
        (4, 61) +- (0, 2.2)
        (5, 63) +- (0, 2.0)
        (6, 64) +- (0, 1.5)
        (7, 65) +- (0, 1.6)
        (8, 67) +- (0, 1.3)
        (9, 69) +- (0, 1.4)
        (10, 72) +- (0, 1.3)
      };
    \end{axis}
  \end{tikzpicture}
  \caption{\last{Win \% of the \sys pipeline over varying numbers of refinement iterations. The win rate reflects how often the \sys pipeline is preferred over a pipeline with only a refiner, as evaluated by GPT-4. Each experiment was repeated three times.}}
  \label{fig:iteration_no_analysis}
\end{figure}
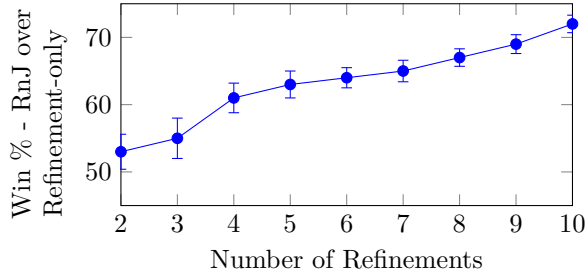

\last{To assess the importance of including a judge in the pipeline, we compared a pipeline with only a refiner to \sys (incorporates both a refiner and a judge). We evaluated the performance of both pipelines across various iteration counts, as shown in Figure~\ref{fig:iteration_no_analysis}. With fewer iterations, the win rate was around 50\%, suggesting that the GPT-4 judge made random selections between the outputs of the refiner-only pipeline and \sys. This outcome is expected, as the LLM-Judge's effectiveness was limited when the refiner performs only one or two iterations. However, as the number of iterations increased, the refiner-only pipeline continued refining without quality checks, while \sys terminated refinements when the quality decreased. Consequently, \sys achieved a win rate of up to 72.5\%, demonstrating that continuous refinements did not guarantee improved quality without a judge.}

\new{We also compared pairwise win rates across refinement iterations for two pipeline variants: one having only the refiner and another having both the refiner and judge (\sys). As shown in Figure~\ref{fig:matrix} in Appendix, the LLM-refiner only pipeline demonstrated inconsistent quality improvements, with several later-stage answers (e.g., $Ans_5$) performing worse than earlier ones (e.g., $Ans_2$), suggesting unstable refinement behavior. In contrast, \sys pipeline exhibited a clear and consistent trajectory of improvement where later answers almost always outperformed earlier ones. For example, $Ans_5$ wined against $Ans_0$ 100\% of the time and wined against $Ans_2$ in 81\% of the time which was only 33\% for the pipeline that only has refiner. These results highlighted that integrating a judgment mechanism not only prevents regression but also ensures that \textbf{each refinement iteration leads to a measurable quality gain.}}

\subsection{Robustness to Noisy Data}
Since \sys was designed to improve low-quality responses ($Ans_0$) obtained from publicly available datasets, it was essential to evaluate whether the pipeline remains effective when the input data is noisy or flawed. In practical applications, online datasets may include inaccurate, verbose, biased, or misleading answers. A robust refinement pipeline should be capable of correcting these flaws rather than perpetuating or amplifying them. 

To analyze this, we tested \sys on intentionally degraded datasets that simulate real-world imperfections. We defined an answer as ``bad" if it exhibited \new{inaccuracy which contained factual errors, verbosity which included long or redundant content, unhelpful where the answer failed to directly address the user's query, or misleading which reflected commonly held false beliefs.}

We constructed our noisy dataset using a combination of public datasets and synthetically generated examples via GPT-4, specifically prompted to produce flawed outputs across the above dimensions, summarized in Table~\ref{tab:dataset_summary}.

To evaluate the robustness, we applied \sys to these noisy inputs and compared its outputs using targeted metrics. \last{We analyzed accuracy improvements from acronym generation to a given title, reduction in verbosity from measuring the average token length, and helpfulness and gains in helpfulness and truthfulness. These gains were assessed by GPT-4, which scored the answers out of 100, allowing us to measure the change in scores.}

\begin{table}[h]
    \centering
    \resizebox{0.6\columnwidth}{!}{%
        \begin{tabular}{lccc}
            \toprule
            \textbf{Dataset} & \textbf{Target} & \textbf{Evaluation} & \textbf{Results} \\
            \midrule
            Acronym & Accuracy & Binary accuracy & +85\% \\
            UltraChat & Verbosity & Length of token & -52\% \\
            OpenAssistant & Helpfulness & LLM-as-a-judge score (0-100) & +66\% \\
            TruthfulQA & Truthfulness & LLM-as-a-judge score (0-100) & +92\% \\
            \bottomrule
        \end{tabular}
    }
    \caption{\sys's performance on noisy data.}
    \label{tab:result_robustness}
\end{table}

Table~\ref{tab:result_robustness} summarizes the pipeline's performance across these noisy conditions. The results showed that \sys is consistently able to improve answer quality, even when the initial input is significantly flawed. On acronym-based tasks, binary accuracy improved by approximately 85\%, in summarization tasks, the average length decreased by 52\%, and for dialogue based tasks, GPT-4 judged a 92\% improvement in truthfulness and 66\% decrease in unhelpful content. These findings showed that \sys not only enhances standard datasets but also offers robust correction capabilities for noisy, and imperfect data.

\subsection{Robustness of LLM-Judge}
Since \sys relies on an LLM-based judge to determine whether one answer is better than another, it is critical to understand the robustness and internal consistency of this judge. A consistent judge should reliably prefer the same choice over time, despite the choices being randomized. Hence, the LLM-judge in \sys should prefer higher-quality responses over earlier, lower-quality ones in a refinement chain. 

To test this, we conducted a self-consistency experiment on the LLM-judge. We evaluated the judge's decision-making by repeatedly presenting it with pairwise comparison between successive answers from the same refinement trajectory -i.e. comparisons between $Ans_i$ and $Ans_{i+1}$. Because each $Ans_{i+1}$ in the chain was generated and previously judged superior to $Ans_i$ during the \sys process, the LLM judge should ideally continue to prefer $Ans_{i+1}$ in subsequent evaluations.

To test robustness of the judge, we randomized the position of $Ans_i$ and $Ans_{i+1}$ across all pairs in our curated data for ten times per pair to ensure that answer quality- not position- was basis for preference. We also tested the judge's reliability in distinguishing valid answers from rejected ones. Specifically, we compared $Ans_n$ (the final accepted answer) with $Ans_(n+1)$ (the rejected one, discarded for failing to improve). Ideally, the judge should continue to favor $Ans_n$ over $Ans_{n+1}$. 

We observed that the judge shows strong agreement early in refinement chain, such as 100\% agreement for $Ans_0$ vs $Ans_1$. However, there was a gradual decline in consistency \last{(from 100\% to 80\% within the first four refinements)} as the refinement chain progress. Near 50\% agreement between $Ans_n$ and $Ans_{n+1}$ was reached, indicating random choice during termination of the pipeline. This pattern suggested that while the LLM-judge is highly consistent in early refinement stage, its confidence diminishes towards the termination point. When differences between successive answers become marginal, the judge appeared to struggle in distinguishing when to end the pipeline, effectively making a guess about whether further refinement was necessary.  

\subsection{Fine-tuning Results}

\new{We fine-tuned Llama 3.1-8B and Llama 3.3-70B models on the original public dataset and \sys enhanced datasets. We then generated responses for AlpacaEval and MT-Bench prompts.}

We first compared three different models. The first was the pre-trained model without any additional fine-tuning. This served as a control to measure the impact of fine-tuning on model performance. The second was the model that is fine-tuned using the original dataset. This step aimed to enhance the model's performance by exposing it to more task-specific data. The last was \sys dataset tuned model which was an improved version of the original dataset. \new{We compared the models using multiple-LLM judges (GPT-4, Gemini, Llama 3.1-70B, and Claude). Table~\ref{tab:judge_comparison} shows these results which indicated a strong judge preference for models fine-tuned on the \sys enhanced dataset compared to original dataset tuned model.}

\new{The Llama 3.3-70B model as a judge demonstrated consistently higher win-rates, reaching up to 80.1\% when tuned-model was also Llama 3.3-70B. The highest win-rate in Llama 3.1-8B tuned model was also observed in Llama 3.3-70B as a judge which achieved above 70\% win-rate.}

\begin{table}[h]
    \small
    \centering
     \begin{tabular}{lcc}
        \toprule
        \textbf{Judge Model} & \textbf{L3-8B Win-Rate} & \textbf{L3-70B Win-Rate} \\
        \midrule
        GPT-4 & 61.6\% & 68.5\% \\
        Gemini & 64.2\% & 69.2\% \\
        Llama 3.3-70B & 69.3\% & 72.1\% \\
        Claude & 62.1\% & 70.9\% \\
        \bottomrule
    \end{tabular}
    \caption{Win-rates between Llama 3.1-8B and Llama 3.3-70B models. The models fine-tuned on the original TULU dataset and the \sys-curated dataset are evaluated by various LLM judges. The win-rate is reported as the percentage of queries where the LLM judge prefers the output from the \sys-incorporated model.}
    \vspace{-5pt}
    \label{tab:judge_comparison}
\end{table}

\begin{table}[h]
    \small
    \centering
    \small
    \begin{tabular}{lcccc}
        \toprule
        \textbf{Model} & \textbf{Tuned Dataset} & \textbf{AlpacaEval} & \textbf{AlpacaEval 2.0} & \textbf{MT-Bench} \\
        \midrule
        \multirow{3}{*}{Llama 3.1-8B} 
            & -     & 70.5 & 24.9 & 6.3 \\
            & TULU  & 79.3 & 34.8 & 6.9 \\
            & \textbf{TULU with \sys}  & \textbf{84.8} & \textbf{37.5} & \textbf{7.5} \\
        \midrule
        \multirow{3}{*}{Llama 3.3-70B} 
            & -     & 85.1 & 39.4 & 8.2 \\
            & TULU  & 88.2 & 51.7 & 8.4 \\
            & \textbf{TULU with \sys}  & \textbf{90.5} & \textbf{54.3} & \textbf{8.6} \\
        \bottomrule
    \end{tabular}
    \caption{Fine-tuning results on different benchmarks.}
    \vspace{-5pt}
    \label{tab:benchmark_comparisons}
\end{table}

\new{Then, we evaluated the performance of fine-tuned models on established benchmarks AlpacaEval, AlpacaEval 2.0, and MT-Bench. The details on the evaluation methodology including the reference and judge models is given in  Table~\ref{tab:evaluation_setup} in Appendix.}

\new{Table~\ref{tab:benchmark_comparisons} presents the win-rates for Llama 3.1-8B and Llama 3.3-70B models fine-tuned using the \sys curated dataset, compared against models fine-tuned on the original TULU dataset and baseline (not fine-tuned) models. These results clearly demonstrate the advantage of applying \sys pipeline to the dataset. For the Llama 3.1-8B model, using the pipeline in fine-tuning improved win-rate of 84.8\% for AlpacaEval compared to 79.3\% with the original TULU dataset. Similarly, performance of AlpacaEval 2.0 and MT-Bench also improved. For the Llama 3.3-70B model, the AlpacaEval performance improved from 88.2\% to 91.8\%, AlpacaEval 2.0 increased from 51.7\% to 53.0\% and MT-Bench resulted in an increase from 8.4 to 8.6 average score. These results consistently demonstrate the effectiveness of \sys pipeline, and showcases meaningful improvements in dataset quality which translates to fine-tuned models' performance results.}

\section{Conclusion}

In this work, we address a crucial aspect of fine-tuning large language models (LLMs): the creation of high-quality training datasets. We introduce the \sys pipeline, which leverages the self-refinement and judgment capabilities of LLMs to automatically generate datasets of preference chains. Our results demonstrate that the \sys pipeline outperforms traditional methods \last{with up to 89\% win rate}, such as generating multiple answers and selecting the best one, as well as zero-shot answer generation. Furthermore, our pipeline surpasses approaches where multiple LLMs generate answers to a single query and the best is chosen, and continuous refinement without judgment.

Our experiments on LLM judge consistency revealed that continuous refinements and judgments lead to consistent choices by the LLM, with improvements that are clear rather than vague. Additionally, the pipeline proved robust to noisy inputs, and models fine-tuned on the \sys-curated dataset demonstrated that this method can enhance LLM performance and tailor models to specific tasks. \last{With these findings, the \sys pipeline emerges as a viable approach for curating preference datasets for fine-tuning LLMs. The pipeline’s prompts can be customized to suit a wide range of scenarios targeting specific goals, such as making LLM responses more engaging by adding friendliness or chattiness as criteria in the refinement and judgment prompts. This enables any dataset to be improved through the pipeline without requiring human annotation. Importantly, these results highlight the potential of the \sys pipeline as a foundation for future research beyond the domains tested in this work.}

\subsection{Limitations}

While our pipeline remarkably enhances the datasets that lead to better fine-tuning results, one limitation of \sys is that the LLM-judge exhibits highly consistent choices \last{(starting from 100\%)} during the early refinement stages, but this consistency diminishes as the process approaches termination. \last{This suggests that when deciding whether to end the refinement chain, the LLM judge’s binary decision often becomes more of a guess, probably due to the marginal improvements in refinements decreasing. To address this,  multi-judge voting can be done. Additionally, as the process approaches termination, using a different LLM model for further refinements can introduce a fresh perspective, potentially leading to more substantial changes in the answers which may motivate the judge to continue with the refinements, and may also reduce the bias resulting from using a single LLM as a refiner and judge.} 

\subsection{Ethics}

\last{The \sys pipeline's automation of dataset generation and refinement raises ethical concerns, particularly regarding the amplification of biases inherent in LLMs. It is crucial to implement mechanisms for bias detection and mitigation to prevent the reinforcement of harmful stereotypes or misinformation. Additionally, fine-tuning LLMs for specific tasks, such as enhancing engagement, must be approached cautiously to avoid manipulative practices. Transparency in how models are fine-tuned and the criteria used in the refinement process should be maintained to uphold user trust and accountability. As the \sys pipeline can be adapted to various domains, it is essential to ensure that its application aligns with ethical guidelines and legal standards specific to each field.}

\clearpage
\newpage
\bibliographystyle{assets/plainnat}
\bibliography{custom}

\clearpage
\newpage
\beginappendix

\startcontents[sections]
\printcontents[sections]{l}{1}{\setcounter{tocdepth}{2}}
\newpage

\section{Broader Related Work}
\label{app:related}

\last{We compare our framework with prior approaches and analyze the differences and our contributions. We specifically compare with preference-based fine-tuning, synthetic preference generation, iterative self-refinement pipelines, LLMs as fine-grained judges, and iterative label refinement approaches. The comparisons are summarized in Table~\ref{tab:comparison_summary}.}

\noindent{\textbf{Comparison with preference-based fine-tuning approaches.}} 
\last{Alignment methods such as Reinforcement Learning from Human Feedback (RLHF) and Direct Preference Optimization (DPO) depend heavily on large-scale human annotations to train a reward model and fine-tune the language model accordingly. However, the collection of such annotations is costly and does not scale well. To address this limitation, recent approaches have explored leveraging large language models (LLMs) to self-generate preference data with minimal human involvement. For example, Selfee~\cite{wang2023aligning} proposes such a method.}

\last{Similarly, another recent work introduces a technique where the language model itself generates rewards for its own outputs during training~\cite{yuan2024self}. This approach uses the model as a judge to evaluate the quality of its responses, effectively creating a reward model from the same model that produces the outputs. This self-rewarding mechanism enables the model to iteratively improve both its instruction-following and reward modeling capabilities without relying on external human feedback or a separate reward model.}

\last{Compared to these works, \sys is about creating datasets of increasing quality answers. Rather than assigning explicit reward scores or labels, our approach uses an iterative refinement-and-judgment cycle in which each refinement step generates progressibely better outputs verified by the same LLM serving as both refiner and judge. Thus, \sys systematically constructs chains of preference-labeled data, providing significant improvements in dataset quality while eliminating reliance on human annotations or seperate reward models.}

\noindent{\textbf{Iterative self-refinement pipelines.}}
\last{Iterative self-refinement methods, such as Self-Refine~\cite{madaan2023self} focus primarily on improving model outputs through internal feedback loops during inference without generating structure training datasets. Furthermore, other recent frameworks like Evolve~\cite{} explore improve self-refinement capabilities of LLMs though iterative preference optimization. Evolve alternates between inference and training phases where model continously produces refined answers which are then filtered using an external reward model. The most ideal answer is used to generate a preference-based dataset used to fine-tune the model used to create refinements.}

\last{In contrast, the \sys pipeline integrates a refiner and a judge that work in alignment at every iteration. At each step, the LLM explicitly evaluates whether the refined answer is an improvement, rather than blindly generating successive refinements without validation. This approach not only reduces computational overhead by eliminating the need for a separate reward model but also ensures that every retained refinement is of higher quality, as verified by the judge. Consequently, unnecessary refinements are avoided, further optimizing computational efficiency while maintaining answer quality.}

\begin{table}[ht]
\centering
\begin{adjustbox}{max width=\columnwidth}
\scriptsize
\begin{tabular}{l c p{3cm} p{3cm} p{3cm} c}
\hline
\textbf{Method} & \textbf{Human-Free} & \textbf{Data Generation Method} & \textbf{Judgment Mechanism} & \textbf{Dataset Improvement Process} & \textbf{Stopping Criterion} \\
\hline
RLHF & No & Human annotation & Human-labeled reward model & Fixed labels (no iterative improvement) & No \\
DPO & No & Human annotation & No & Yes & No \\
Selfee & No & LLM-generated synthetic labels & External (LLM logits-based scoring) & Weak & No \\
Self-Refine & Yes & Single model refinement loop & No & No dataset generation, only improves inference outputs & No \\
EVOLVE & No & Iterative refinement with external scoring & External (separate reward model) & Progressive refinement with external filtering & External \\
\textbf{\sys} & \textbf{Yes} & \textbf{Iterative refinement verified by judge} & \textbf{Integrated (single LLM as refiner and judge)} & \textbf{Preference labeled data generation through iterative verified refinements} & \textbf{Yes} \\
\hline
\end{tabular}
\end{adjustbox}
\normalsize
\caption{Comparison summary of \sys and related approaches}
\label{tab:comparison_summary}
\end{table}

\newpage
\clearpage

\section{Illustration of Only Refinement Pipeline}
\begin{figure}[H]
    \centering
    \includegraphics[width=0.5\linewidth]{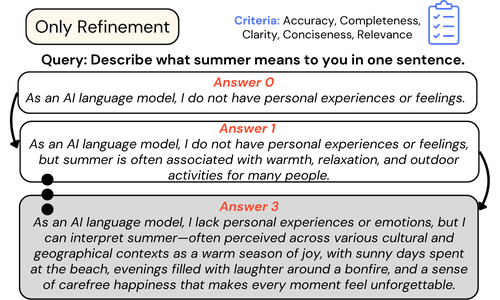}
    \caption{Illustrative example of iterative refinement pipeline without a stopping condition.}
    \label{fig:refinement_only_ex}
\end{figure}

\new{Prior work has shown how LLMs can be used to refine their own answers~\cite{madaan2023self}. In Figure~\ref{fig:refinement_only_ex} we explore a similar pipeline where only continous refinements are made without incorperating a judge. The prompt asks for one sentence description of the summer. Starting from an initial model output, each subsequent answer is generated by prompting the model to improve upon the previous one.}

\new{As can be seen from Figure~\ref{fig:refinement_only_ex}, without a judge, iterative refinement increased verbosity. Since the refinements did not terminate, this pipeline would continue making refinements due to the lack of a stopping condition. However, by incorporation a judgment step at each iteration, \sys method ensured refinements are both substantive and aligned with quality criteria, and terminated when further refinements were no longer beneficial.}

\begin{figure}
  \centering
  \begin{subfigure}[b]{\columnwidth}
    \centering
    \includegraphics[width=0.6\columnwidth]{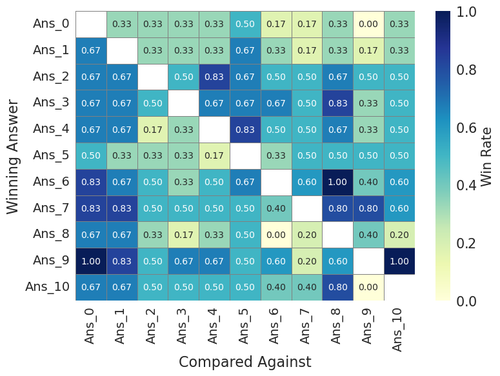} 
    \caption{Only LLM-Refine pipeline.}
    \label{fig:sr_matrix}
  \end{subfigure}
  
  \vspace{1em} %
  
  \begin{subfigure}[b]{\columnwidth}
    \centering
    \includegraphics[width=0.6\columnwidth]{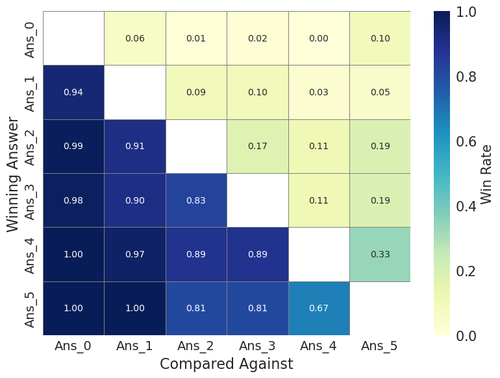} 
    \caption{\sys pipeline.}
    \label{fig:rj_matrix}
  \end{subfigure}
  
  \caption{\new{Pairwise win rate comparison across iterations for two pipelines: (a) LLM-Refine only, proposed in Self-Refine~\cite{madaan2023self}, and (b) \sys. For the \sys pipeline, the matrix spans up to $Ans_5$ due to the decreasing number of instances where the refinement count exceeds 5.}}
  \label{fig:matrix}
\end{figure}

\last{These findings are further supported by the pairwise win rate comparisons showcased in Figure~\ref{fig:matrix}. When using only the refinement pipeline without a judging component (Figure~\ref{fig:sr_matrix}), later answers were not consistently better than earlier ones. For example, $Ans_9$ outperformed $Ans_6$ only 40\% of the time, indicating that the quality of answers degraded during the three refinement steps from $Ans_6$ to $Ans_9$. This degradation occurred because, in the absence of a judge, lower-quality refinements were retained.}

\last{In contrast, incorporating a judge into the pipeline led to a more favorable pattern, as shown in the pairwise comparison matrix  in Figure~\ref{fig:rj_matrix}. Here, the win rates increased in a downward direction and decreased horizontally to the right, reflecting a consistent improvement in answer quality across iterations. For instance, $Ans_1$ was preferred over  $Ans_0$ 94\% of the time, and $Ans_4$ was favored over $Ans_0$ in 100\% of the comparisons. This demonstrates the effectiveness of judgment-based selection in promoting higher-quality refinements.} 

\newpage
\clearpage

\section{\sys Prompts}

\begin{figure}[H]
    \centering
    \includegraphics[width=0.6\linewidth]{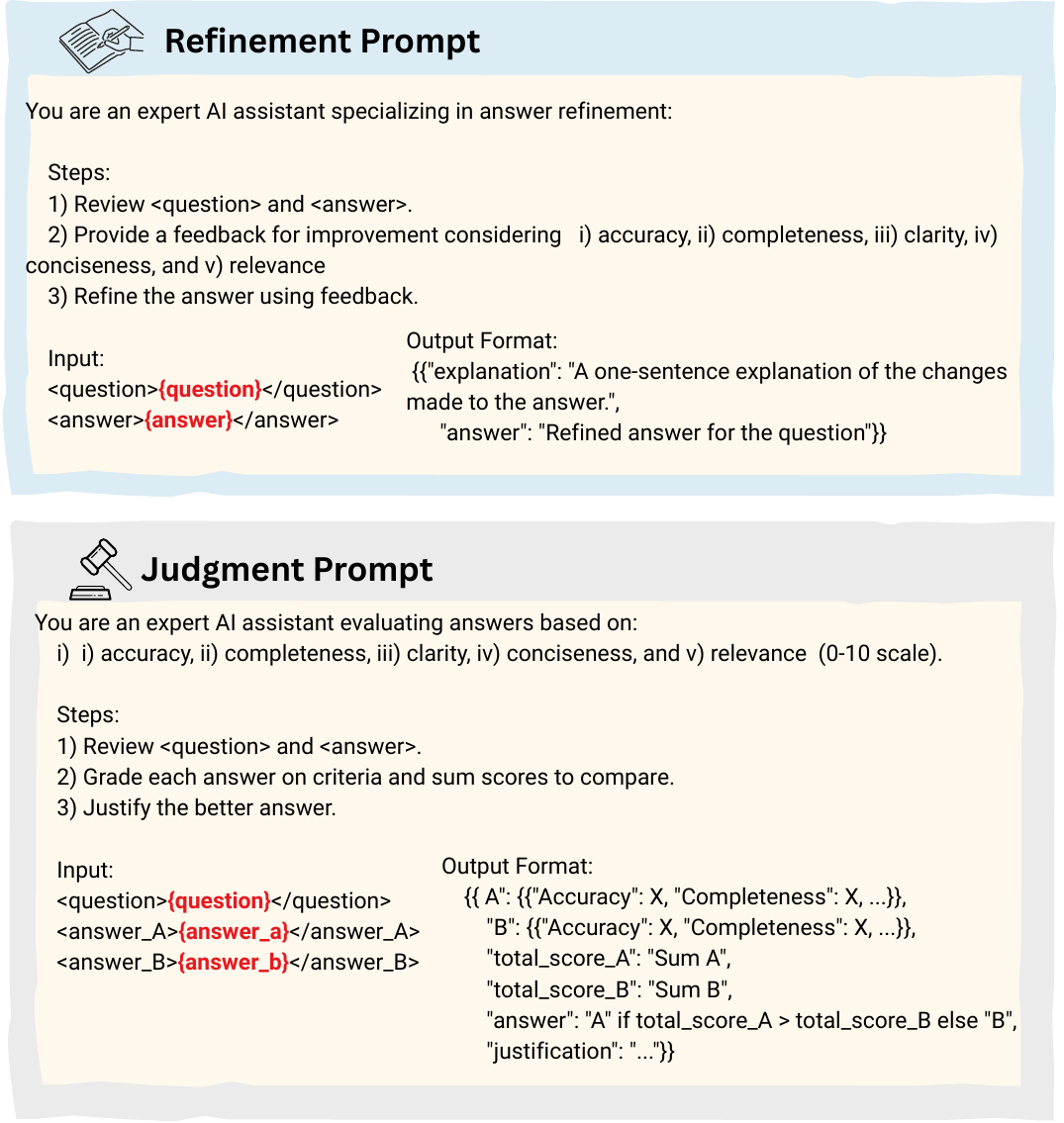}
    \caption{\last{Prompts for \sys where sequential processes: feedback generation, refinement, and judgment.}}
    \label{fig:prompts}
\end{figure}

\last{Figure~\ref{fig:prompts} illustrates the prompts used in \sys.}

\subsection{LLM-Refiner:} 
\new{To refine the initial answer, we first ask the LLM to generate a feedback on the answer. For this, we give the question and the answer that is wished to be refined. The LLM uses a set of criteria to base their feedback on.}
\new{After the feedback is generated, it is used to compute the refined answer.}

\subsection{LLM-Judge:}
\new{We leverage the capabilities of LLMs to assess and provide reasoning for complex tasks. Given a question and two possible responses,  LLM is prompted to grade each response based on a predefined set of criteria. This prompt is designed to be generalizable across various tasks without modification, although adjustments can be made depending on the dataset or task if desired.  LLM assigns scores to each criterion for both responses and sums these scores to determine which response is ranked higher. Based on this evaluation, LLM selects a preferred answer and provides a concise justification for its choice. } 

\new{It is important to note that during the evaluation process, the positions of the answers are systematically swapped. Specifically, the locations of $Ans_{i}$ and $Ans_{i+1}$ are alternated to ensure that the subsequent answer $(i+1)$ is not consistently presented as $answer_A$. This procedure helps mitigate the positional bias in LLM's judgments.}

\newpage
\clearpage

\section{Computational Cost of \sys}
\label{sec:gpu_appendix}

\begin{figure}[h]
  \centering
  \begin{tikzpicture}
\centering
\begin{axis}[
    ybar,
    symbolic x coords={1,2,3,4,5,6},
    xtick=data,
    xlabel={Number of answers},
    ylabel={Number of samples},
    ymin=0,
    ymax=35000,
    bar width=15pt,
    nodes near coords,
    nodes near coords align={vertical},
    width=9cm,
    height=5cm,
    grid=major,
]
\addplot coordinates {
    (1,28080)
    (2,27456)
    (3,14377)
    (4,5257)
    (5,1925)
    (6,906)
};
\end{axis}
\end{tikzpicture}
\caption{\new{The number of refined answers obtained from \sys pipeline to TULU dataset. The total number of samples is 78k.}}
\label{fig:refinement_counts_tulu}
\end{figure}
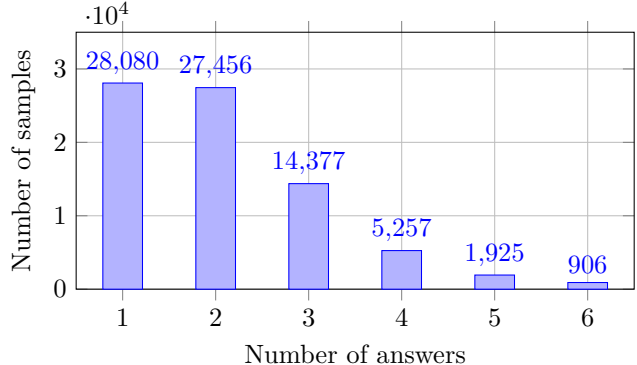

\new{%
The pipeline was run on TULU dataset~\cite{lambert2024tulu3} where the initial dataset contained 78k samples. The number of refinements count for the sample sizes are given in Figure~\ref{fig:refinement_counts_tulu}. Majority of the queries had only one or two refinements done by the pipeline.}

\section{Evaluation Details}
\new{We primarily assess our models using two of the most popular chat-based evaluation benchmarks AlpacEval2~\cite{alpaca_eval} and MT-Bench~\cite{zheng2023judging}. These benchmarks have been widely adopted by the community due to their versatile conversational capabilities across many domains. AlpacaEval 2 consists of 805 questions from 5 datasets, and MT-Bench covers 8 categories with 80 questions. We ue GPT-4-Preview as the baseline model in AlpacEval2.0, which is the model our results are compared against. We report the length controlled win rate against GPT-4-Preview using GPT-4 as a judge where the temperature is set to zero. For MT-Bench, we report the average MT-Bench score through GPT-4 as the judge model. The evaluation details for the benchmarks are summarized in Table~\ref{tab:evaluation_setup}.} %

\begin{table}[H]
    \centering
     \begin{tabular}{>{\bfseries}lcccc}
        \toprule
        Benchmark & Baseline Model & Judge Model & Scoring Type & Metric \\
        \midrule
        AlpacaEval & Davinci003 & GPT-4-Turbo & Pairwise comparison & Win rate\\
        AlpacaEval 2 & GPT-4-Preview & GPT-4-Turbo & Pairwise comparison & LC Win rate \\
        MT-Bench & - & GPT-4-Turbo & Answer grading & 1-10 score \\
        \bottomrule
    \end{tabular}
    \caption{Evaluation details for SFT results.}
    \label{tab:evaluation_setup}
\end{table}

\newpage
\clearpage

\end{document}